\documentclass[
]{ceurart}

\usepackage{listings}
\usepackage{tabularx}
\usepackage{booktabs}

\begin{document}

\copyrightyear{2025}
\copyrightclause{Copyright for this paper by its authors.
  Use permitted under Creative Commons License Attribution 4.0
  International (CC BY 4.0).}

%\conference{Woodstock'22: Symposium on the irreproducible science,
%  June 07--11, 2022, Woodstock, NY}
\conference{Nara 25: The 20th International Workshop on Ontology Matching,
November 02--03, 2025, Nara, Japan}
% The 20th International Workshop on Ontology Matching
%collocated with the 24th International Semantic Web Conference ISWC-2025,
% November 03--04, 2025, Nara, Japan
%November 2nd or 3rd, 2025, Nara, Japan

\title{Adaptive and Multi-Source Entity Matching for Name Standardization of Astronomical Observation Facilities}

%\tnotetext[1]{You can use this document as the template for preparing your
%  publication. We recommend using the latest version of the ceurart style.}

\author[1]{Liza Fretel}[%
orcid=0009-0001-4600-9954,
email=liza.fretel@obspm.fr,
url=https://github.com/Sazuna,
]
\cormark[1]
\fnmark[1]
\address[1]{Paris Observatory, Pl. Jules Janssen, 92190, Meudon, France}

\author[1]{Baptiste Cecconi}[%
orcid=0000-0001-7915-5571,
email=baptiste.cecconi@obspm.fr,
]

%\cortext[1]{Corresponding author.}
%\fntext[1]{These authors contributed equally.}

\author[1]{Laura Debisschop}[%
orcid=0000-0003-4688-6575,
email=laura.debisschop@obspm.fr,
]

\begin{abstract}
  This ongoing work focuses on the development of a methodology for generating a multi-source mapping of astronomical observation facilities. To compare two entities, we compute scores with adaptable criteria and Natural Language Processing (NLP) techniques (Bag-of-Words approaches, sequential approaches, and surface approaches) to map entities extracted from eight semantic artifacts, including Wikidata and astronomy-oriented resources. We utilize every property available, such as labels, definitions, descriptions, external identifiers, and more domain-specific properties, such as the observation wavebands, spacecraft launch dates, funding agencies, etc. Finally, we use a Large Language Model (LLM) to accept or reject a mapping suggestion and provide a justification, ensuring the plausibility and FAIRness of the validated synonym pairs. The resulting mapping is composed of multi-source synonym sets providing only one standardized label per entity. Those mappings will be used to feed our Name Resolver API and will be integrated into the International Virtual Observatory Alliance (IVOA) Vocabularies and the OntoPortal-Astro platform.
\end{abstract}


\maketitle

\section{Introduction}

  \subsection{Context}
  Astrophysics brings together a wide range of dynamic communities: heliophysicists, planetary scientists, cosmologists, data scientists, instrument engineers, etc. These diverse experts collaborate across disciplines to tackle complex questions about the Universe. % --- from the behavior of stellar winds and planetary magnetospheres to the structure of galaxies and the evolution of cosmic large-scale structures. 
   The astrophysics community has also been pioneering open-science with the early inception of the so-called \emph{Virtual Observatory} \cite{genova_ivoa_cds_2000,arviset2018vopowerfultoolglobal} in the early 2000's, which defines open and interoperable data and access standards for astronomy, that are maintained and developed by the IVOA \cite{arviset2012}. Similar initiatives also exist for heliophysics (International Heliophysics Data Environment Alliance, \href{https://ihdea.net}{IHDEA}) \cite{roberts2018spase,fung2023spase} and planetary sciences (International Planetary Data Alliance, \href{https://planetarydata.org}{IPDA}) \cite{slavney2007ipda}.
  This interdisciplinary synergy is further amplified by the rapid development of computational tools, and globally shared data standards and open-access practices, such as those developed within the Research Data Alliance \cite{showstack2012rda}.
  
  As data becomes increasingly interconnected, there is a growing need to ensure its interoperability. Specifically, the data discovery part of the science workflow is greatly facilitated when the data providers are using metadata with the same schema and terms from the same vocabularies. However, research institutions often do not use consistent naming conventions, particularly when referring to astronomy observation facilities (equivalent to the \emph{platform} concept defined in OGC SensorML\footnote{Open Geospatial Consortium Sensor Model Language: \url{https://docs.ogc.org/is/23-000/23-000.html}}),
%  (\textit{Telescopes, Observatories, Spacecraft, Airborne platforms}, and \textit{Investigations}):
some refer to a telescope by its diameter and/or location, others by its nickname, and many variations exist that combine both. Additionally, historical labels (like "Mariner 11" for "Voyager 1") for the same facility can introduce further ambiguity, requiring a deeper understanding of the definitions and contexts around these entities.
  
  The main objective of this work is to propose a method to align multiple lists of astronomical observation facilities, standardize their aliases, and suggest a unique preferred label per physical entity. The source code of the functionalities described in this paper are available on GitHub \cite{fretel_2025_17199128}. The data can be re-downloaded by running \textit{update.py}. A snapshot of the cached pages on 2025/09/04 is available on Zenodo \cite{fretel_2025_17078681}.

  \subsection{Applications}
  \label{sec:applications}
  
  The primary goal of this work is to enable smooth data discovery across data providers, in the scope of the several astronomy open data ecosystems (including the IVOA, IHDEA and IPDA). We thus need to propose, firstly, a vocabulary of commonly agreed terms, that shall be used to expose metadata in search interfaces; and secondly,   
  a name resolver that suggests matching entities 
  from the adopted vocabulary. %along with their recommended labels for any given input string. 
  
  The implementation of the name resolver relies on an Elasticsearch API, powered by a JSON dictionary (example in Appendix \ref{appendix:json}) that includes the preferred labels of observation facilities along with their aliases. The name resolver will have two interfaces: a \emph{resolve} endpoint, which takes any input string and provides an ordered list of matching terms; and an \emph{aliases} endpoint, which takes a term from the vocabulary and returns the list of known aliases. That second endpoint is also useful for expanding data discovery to databases that have not adopted the proposed vocabulary yet. In order to implement refined search and discovery, we also include a search by meronymy relation so that the data discovery is not restricted to the granularity choice of the data provider (for instance, a user searching for observations conducted by the \emph{Voyager} space mission, shall also find the %entries labelled with any of the
  \emph{Voyager 1} and \emph{Voyager 2} spacecraft). 
  % At the time of writing, the current implementation is based on a preliminary version of the vocabulary, and is only accessible through direct web queries, thus we .  
  % This tool will support researchers in standardizing the name of observation facilities
  %=> pas certain 
  
  In parallel to this work, we develop OntoPortal-Astro\footnote{\url{https://ontoportal-astro.eu}} \cite{cecconi2025ontoportalastrosemanticartefactcatalogue} to share astronomy ontologies while ensuring the FAIRness of their content. It aligns with the OntoPortal Alliance, a broader initiative to build portals for domain-specific ontologies, such as EarthPortal \url{https://earthportal.eu/} and AgroPortal \url{https://agroportal.lirmm.fr/}. The ontology produced by our multi-source mapping will later be shared on OntoPortal-Astro, allowing the users to access statements from diverse resources and to track the statements' origin.

  Finally, we also output an observation facilities list following the CSV format defined by the IVOA \cite{demleitner2023ivoavcabularies}, so that it can be processed and listed on the IVOA vocabulary page\footnote{\url{https://ivoa.net/rdf}}. This CSV file lists every physical entity and the meronymy relations between them. 

\subsection{Related works}
   The previous version of our algorithm \cite{Cecconi2023ObsFacilityWikidata} used to set Wikidata as a \textit{pivot} ontology, trying to map every other ontology to it, but our attempts were unsuccessful, due to the non-exhaustiveness of Wikidata. That is why we introduced the mapping strategy, that permits to link resources in any order and on any criterion. The necessity of a multi-strategy approach to map plural and heterogeneous entity collections was highlighted in the original RiMOM paper \cite{riMOM2009multistrategyOntologyAlignmentFramework}.
   
   % OntoAligner \cite{Babaei_Giglou_OntoAligner_A_Comprehensive_2025} proposes to encode data and retrieve the nearest entities based on different encoding models such as a TF-IDF encoder, Transformers like the \textit{all-MiniLM-L6-v2}, few-shot models, etc.
   % Furthermore, we leveraged several commonly used semantic scores, ... such as the Tf-Idf that were exploited in the ...  paper.
   In our work, we employ an LLM to validate or invalidate a candidate pair, which is composed by two semantically close entities. This method was exploited by LLM-Align  \cite{chen2024llmalignutilizinglargelanguage}, a framework that embeds entities and selects the k-nearest target entities before asking an LLM to decide which target entity matches the source entity. % In both cases, the LLM plays a key role in deciding the final mapping.

\section{Data updating}
\label{sec:updating}

  \begin{figure}
   \centering
   \includegraphics[width=16cm]{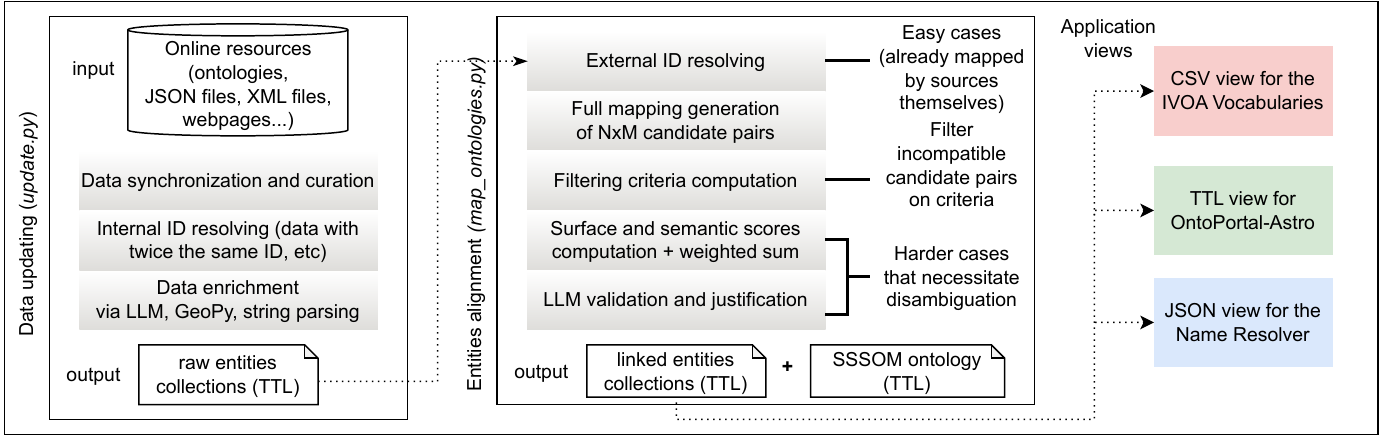}
   \caption{Data processing pipeline. During data updating (\textbf{left}), described in section \ref{sec:updating}, we collect observation facilities' records and save them in turtle files. Those files are used as inputs of the entity alignment steps (\textbf{right}), described in sections \ref{sec:mapping}, \ref{sec:scores} and \ref{sec:llm_validation}. This outputs a linked ontology containing all entities from each list with their matching relations (\textit{SKOS:exactMatch}), along with an associated SSSOM ontology. The application views can be generated from the linked ontology. Their purposes are explained in the subsection \ref{sec:applications}.
   }
   \label{fig:process}
 \end{figure}

  \subsection{Data sources, processing and updates}
 
  %Through this research, we are looking for an optimized way to align observation facilities from different online resources (web pages, ontologies, lexicon, etc.). 
 Currently, we only processed eight out of 19 identified vocabularies \cite{fretel_2025_15862784}:
 Wikidata \cite{vrandevcic2014wikidata}, \href{https://journals.aas.org/facility-keywords/}{AAS} (American Astronomy Society), \href{https://pds.nasa.gov/data/pds4/context-pds4/}{PDS} (Planetary Data System)%\cite{PDS4ContextProducts}
 , \href{https://www.minorplanetcenter.net/iau/lists/ObsCodesF.html}{IAU-MPC}\footnote{IAU-MPC entities are referred to with an identifier in Wikidata (\href{https://wikidata.org/wiki/Property:P717}{Minor Planet Center ID})} (The International Astronomical Union Minor Planet Center)%\cite{iaumpcCodes}
 , \href{https://naif.jpl.nasa.gov/pub/naif/toolkit_docs/FORTRAN/req/naif_ids.html}{NAIF}%\cite{naif}
 \footnote{NAIF entities are referred to with an identifier in Wikidata (\href{https://www.wikidata.org/wiki/Property:P2956}{NAIF ID}).} (The Navigation and Ancillary Information Facility)
 , \href{https://nssdc.gsfc.nasa.gov/nmc/}{NSSDC}%\cite{nssdca_master_catalog}
 \footnote{NSSDC  entities are referred to with an identifier in Wikidata (\href{https://www.wikidata.org/wiki/Property:P247}{COSPAR ID} and \href{https://www.wikidata.org/wiki/Property:P8913}{NSSDCA ID})} (NASA Space Science Data Coordinated Archive), \href{https://spase-group.github.io/spase-info/index.html}{SPASE} (Space Physics Archive Search and Extract) %\cite{spase-info}
  and \href{https://ssp.imcce.fr/webservices/ssodnet/api/quaero/}{IMCCE} (Celestial Mechanics and Ephemerides Calculus Institute).%Institut de mécanique céleste et de calcul des éphémérides  %\footnote{The Sso name search is powered by LTE'sSsODNet.quaero REST API.}.
  The extracted observation facilities fit into five classes: \textit{Telescope, Observatory, Spacecraft, Airborne platform}, and \textit{Investigation}. Those classes are convenient for the ontology mapping task, but do not embody the complexity of the taxonomy of observation facilities. We might consider adding other classes when needed. % More identified resources were detailed in our IVOA note ``Using Wikidata for an Observation Facility Vocabulary''\cite{Cecconi2023ObsFacilityWikidata}. => Je ne le dis plus puisque j'ai cité mes diapos (et que je re-cite la note au paragraphe juste après)
 
   \subsubsection{Data synchronization and curation}
  
  In the previous version of the data format (detailed in \cite{Cecconi2023ObsFacilityWikidata}), data were not typed and only labels and identifiers were mapped with a threshold on the Levenshtein similarity \cite{levenshtein1966binary}. However, in doing so, the labels did not provide enough information to be properly disambiguated. To counter this, we extracted every field that might add a semantic or numerical value to the entities by enriching them with a class if possible, a description and/or definition, an observation waveband, a launch date for spacecraft, a funding agency, a latitude and longitude for ground-based facilities, etc. A sample of the extracted data is available in Appendix \ref{appendix:data}. 
  
  One of the nice-to-have features mentioned in \cite{Cecconi2023ObsFacilityWikidata} is the automated update feature. We developed a \textit{version manager} that detects any change on an entity whenever \textit{update.py} is called with the \textit{--no\_cache} argument. It updates its \textit{DCTERMS:Modified} value and sets a \textit{Deprecated} flag to entities that no longer exist in their original source.% They are kept to keep track of the facilities' old naming conventions.
  
  After data were synchronized, we fix the identified errors in the data (data curation step).
  
  \subsubsection{Internal ID resolving} 
  
  There are several constraints regarding the URIs management. Since the resulting data will be mainly exploited in the IVOA ecosystem, we have to conform to their specifications, such as the use of the facilities' labels to generate human-readable URIs for historical reasons. This is an issue for sources that have non-unique labels, which we solved by adding keywords or by re-using the source's identifiers as URIs, like in IAU-MPC or PDS, in which an investigation and a spacecraft are distinct records but share the same label. Moreover, we defined a namespace per source (see Appendix \ref{tab:namespaces} for namespaces used in this document) to prevent entities from two resources but with the same label to end up with the same URI. In addition, the NAIF list contains non-unique identifiers for the same or different entities, which required domain experts to be resolved. % Finally, we replace the source's URIs or identifiers by our local ontology URIs.

    %\subsubsection{Version management}
  %One of the nice-to-have features mentioned in \cite{Cecconi2023ObsFacilityWikidata} was the automated updates feature. We developed a \textit{version manager} that detects any change on an entity and will update its \textit{DCTERMS:Modified} value. It sets a \textit{Deprecated} flag to entities that no longer exist in their original source, but we do not want to delete it to keep track of the facilities' old naming conventions.

    \subsubsection{Data enrichment}
  Finally, the data enrichment step consists of using tools to fill missing attributes. As most of the entities' classes can be deduced from their metadata (Wikidata), their URL (PDS) or the web page layout (NAIF), some of the lists (SPASE) do not explicitly classify entities. We prompted an LLM to classify non-explicitly typed resources into our five categories (six with the \textit{Unknown} class for entities that do not belong to any category, like space military facilities). Unfortunately, our best attempt with the model \href{https://huggingface.co/meta-llama/Llama-3.1-8B}{LLama3.1-8B} \cite{touvron2023llamaopenefficientfoundation} only reached 80.20\% accuracy on the classification task, and the results were too sensitive to the prompt. Because we did not want to introduce biases at the early stages of the mapping process, we decided to consider those entities as untyped, therefore we will try to pair them with every entity instead of creating one sub-mapping per type (e.g. a mapping between IAU-MPC and Wikidata's observatories).
  
  Then, we used Geopy (version 2.4.1) to retrieve the ground-based entities' location information such as a detailed address, latitude and longitude from the entities' metadata. The address can benefit the semantic scores because some facilities share their name with their street, city or country.
  
  String parsing is used for example on PDS spacecraft's aliases to extract their NSSDC identifiers and infer their launch years, as well as on AAS facilities to extract their apertures (size of the primary mirror or the input lens of a telescope) and some aliases. An example of data enrichment can be found in \ref{appendix:alt_labels}.

  \section{Mapping strategy}
  \label{sec:mapping}
   
  The mapping strategy is a configuration file that allows the engineer to program a mapping path (see an example in Appendix \ref{appendix:mapping_strat}) on a certain type and on certain scores only, proceeding by pairs of lists. In this way, we iteratively increase the size of the synonym sets, which are made of validated synonym pairs, enhancing the subsequent mapping steps. % As a synonym set grows, more data is added to it
  % Each line of the configuration document plans a mapping between two sets of entities

  The mapping strategy splits the entities alignment problem into three levels of difficulty. Those levels are displayed in the right part of Figure \ref{fig:process}). When being confronted to two sets of entities, we start with the most obvious alignments --- external identifiers (external ID resolving step), linking entities that are already refer to each other. After that, we generate a full mapping with the remaining single entities (second step). In this step, \(N \times M\) candidate pair objects are generated, with \(N\) and \(M\) varying between 561 (PDS) and 10.000 (IMCCE) entities. We then apply discriminant criteria (third step), reducing the complexity depending on the criteria and the lists' features, and we end up with the most complex mappings, which require the computation of surface and semantic scores, followed by LLM validation to disambiguate them (the last two steps).% The discriminant criteria and scores are further explained in the next subsection. %The full mapping generation (second step) were optimized with parallelization, as the generation 

  \section{Discriminant criteria and scores}
  \label{sec:scores}
  
  For each line of the mapping strategy, the algorithm will apply a set of filtering criteria, surface and semantic scores on the candidate pairs. %The scores are divided into two categories, detailed in the next two subsections.
  
  %\begin{itemize}
  %    \item discriminant criteria, that include a label match (accepting score)\footnote{The label match score works well on spacecraft and investigations, as their names do not vary much, but might fail on ambiguous entities like a spacecraft and investigation, or a telescope named only after its lens aperture or host observatory.}, a mismatch in location, dates, classes, aperture size, or identifiers (rejecting scores);
  %    \item surface scores, that consist of comparing labels' forms with Levenshtein similarity, acronym detection, and digits correspondence ratio;
  %    \item finally, semantic similarity scores, which embed entities into vector spaces and compute a cosine similarity between two vectors. We allow the user to set any encoder from the TF-IDF (Bag-of-Words (BoW) approach), a sentence transformers or an LLM (sequential approaches) to encode entities.
  %\end{itemize}
  
  \subsection{Filtering criteria}

  Filtering criteria are elimination rules that will accept or disqualify candidate pairs in prior to any further computation. This has two purposes: reduce the mapping's complexity by removing incompatible pairs; and prevent incoherent decisions during the subsequent steps. It includes a label match (accepting criterion)\footnote{The label match works well on spacecraft and investigations, as their names do not vary much.}%, but fails on ambiguous entities like a spacecraft and investigation, or a telescope named only after its lens aperture or host observatory.}
  , a mismatch in location, date, class, aperture size, or identifier (rejecting criteria). For instance, the ``date'' filter compares and disqualifies two spacecraft with different launch years. Ground-based facilities like telescopes or observatories often have an associated latitude and longitude, allowing a geodesic distance computation. We set the maximum distance between two entities to 4 km to account for rounding tolerance; beyond that, they are considered distinct.

    \subsection{Similarity scores}

  Once this preliminary step is done, only compatible pairs are left. To disambiguate, we apply some linear similarity scores, that take different aspects of the entities into account.
  
  \subsubsection{Surface scores}
  The Levenshtein distance \cite{levenshtein1966binary} is an edition distance between two strings. By applying the following formula, we obtain a similarity score between 0 and 1 for strings \(|\text{s}_1|\) and \(|\text{s}_2|\):
  
    \[
    \textsl{LevensteinSimilarity} = 1 - \frac{\textsl{LevensteinDistance}(|\text{s}_1|, |\text{s}_2|)}{\max(|\text{s}_1|,|\text{s}_2|)}
    \]
  
  For example, it outputs a high score between ``observatory'' and ``observatoire'' (French), therefore it is able to detect slight translation variations or typos. For the digits match score, we extract all numbers from the entity's strings with a regular expression, apply truncation and rounding to match numbers from both entities and compute a matching ratio. The acronym probability score computes the probability of a label to be the acronym of another label.
  
  \subsubsection{Semantic scores}
  The TF-IDF (term-frequency inverse document frequency) is used in many OM frameworks \cite{Guli2013OntologyMatchingTFIDF}. 
  % \[tfidf = tf_ij * log(N/df_i)\]
  \[
\textsl{tf-idf}_{i,j} = \textsl{tf}_{i,j} \times \log\left(\frac{N}{\textsl{df}_i}\right)
\]
    In the context of OM, the term frequency \emph{tf} is how often a token appears in an entity's textual fields; \(N\) is the amount of entities that constitute the corpus; \(df\), the document frequency, is how many times the token appears in the ontology's textual fields. It emphasizes the importance of rarer tokens such as proper nouns, while lowering the impact of recurring tokens like ``mission'', but it does not embed the synonyms and sentences' meaning. To train a TF-IDF encoder, we simulate a reference corpus by extracting all of the textual fields (labels, descriptions and definitions) and filter out English, Spanish and French stop words. After that, we can encode the source and the target entities and compute a cosine similarity between their vectors (A and B):

    \[\cos(\theta) = \frac{\mathbf{A} \cdot \mathbf{B}}{\|\mathbf{A}\| \|\mathbf{B}\|}\]

  To encode entities, we also implemented the sentence transformer's cosine similarity. We chose the \href{
https://huggingface.co/sentence-transformers/all-MiniLM-L6-v2}{all-Mini-LM-L6-v2} model. It was pre-trained on a general and multilingual corpus. After entities are vectorized and saved on the disk, we compare them by computing their cosine similarity.
  
  Some LLMs provide an encoding function that generates embeddings for a given text, allowing entities comparison via the cosine similarity. We tested different LLMs of different sizes: \href{https://huggingface.co/meta-llama/Llama-3.1-8B}{LLama3.1-8B} \cite{touvron2023llamaopenefficientfoundation} (8 billion parameters), \href{https://ollama.com/library/deepseek-v3}{DeepSeek-V3} \cite{deepseekai2025deepseekv3technicalreport} (671 billion parameters, version \verb|deepseek-v3:671b-q4_K_M|) and \href{https://huggingface.co/UniverseTBD/astrollama}{Astrollama} \cite{nguyen2023astrollamaspecializedfoundationmodels} (7 billion parameters, fine-tuned on 300.000 arXiv articles). %We had high hopes for Astrollama as it was fine-tuned on astronomy data, but unfortunately,
  Despise being fine-tuned on astronomy data, it did not outperform DeepSeek and its embeddings were too space-greedy.%\footnote{TODO experiment more with Astrollama/DeepSeek ?}.

\subsubsection{Global score}
 By combining those scores via a weighted sum, we obtain a global score on each candidate pair:
 
  \[
    \textsl{score}(p) = \frac{1}{\sum_{i=1}^{n}{w_i}}\sum_{i=1}^{n}s_i \times{w_i}
  \]
  
  where \(p\) is a candidate pair, \(s_i\) is a score value (for instance the TF-IDF's cosine similarity score) and \(w_i\) is its score's weight, a constant that we fix beforehand (for instance we set 0.5 for the Levenshtein similarity). Unlike in a traditional vector space, which is usually tied to a single unique similarity score, this encompasses both the surface and the semantic aspect of the data.
 
 \section{Iterative validation by LLM prompting}
 \label{sec:llm_validation}
 
 For each pair, starting with the one with the highest global score, we prompt an LLM to accept or reject the pair and justify its decision \cite{peeters2024entitymatchingusinglarge}. In the prompt, both entities are represented by a string containing all of their textual features. We also give the instruction to consider a narrower distinct from its broader entity. % that we hope the LLM will be able to process and compare.
 %Positive matches remove the entities and all of its associated candidate pairs from the pool to fasten the validation process.
  After rejecting a certain amount of pairs in a row, we interrupt the process and output an auxiliary SSSOM ontology \cite{SSSOM2022}, which keeps track of the mappings' decision time, LLM justifications, which scores were the decisive ones, etc., for each positive match. Two SSSOM match examples can be found in Appendix \ref{appendix:sssom}.
 
 %\subsection{LLM validation evaluation}
 To evaluate the LLM validation, we used the AAS and PDS facility lists, that share similar features (aperture, naming convention, etc). We annotated a list of 30 pairs of potential matches with compatible features with a ``same|distinct'' label. This experiment was made using \href{https://ollama.com/library/deepseek-v3}{DeepSeek-V3} \cite{deepseekai2025deepseekv3technicalreport} (671 billion parameters, version \verb|deepseek-v3:671b-q4_K_M|) as the candidate pairs' validator. Out of 30 candidate pairs, none were wrongly reviewed: 19 were true positives and 11 true negatives. % In a future work, we will investigate mappings between other lists and give insights about each scores' relevance to different mappings. % An example of a generated SSSOM is shown in the appendix \ref{appendix:sssom}. % je l'ai déjà dit ça

%data_merger.scorer.date_scorer.compute          11.475896058720537
%data_merger.scorer.distance_scorer.compute              19.378836239455268
%data_merger.scorer.fuzzy_scorer.compute         19.74517043742526
%data_merger.scorer.digit_scorer.compute         28.150079354294576
%data_merger.scorer.tfidf_scorer.compute         1072.9254022172245

%# PDS/AAS (with label_match)
%data_merger.scorer.label_match_scorer.compute           3.2282855925150216
%data_merger.scorer.date_scorer.compute          4.5873049513902515
%data_merger.scorer.fuzzy_scorer.compute         7.208128882106394
%data_merger.scorer.distance_scorer.compute              8.840426769573241
%data_merger.scorer.digit_scorer.compute         10.587980876443908
%data_merger.scorer.tfidf_scorer.compute         449.6707207572181

  \section{Conclusion and perspectives}
  In this work, we have introduced the mapping strategy methodology to perform data alignment of astronomy observation facilities. Our contribution mostly consists of the development of an adaptable mapping strategy combining filtering criteria, surface and semantic scores and the iterative aspect of the mapping, enhancing the semantic similarity capabilities by adding new information as we discover new synonyms.
  
  %two points:
%\begin{itemize}
%\item the development of an adaptable mapping strategy combining discriminant, surface and semantic scores;
%\item the iterative aspect of the mapping, enhancing the semantic similarity detection capabilities by adding new information as we discover new synonyms.
%\end{itemize}

 As we focused on a limited amount of resources, in the future, we would like to align more vocabularies together. Furthermore, we are currently annotating a dataset of candidate pairs with a ``same|distinct'' annotation with our European collaborators for the disambiguation task on observation facilities. It will help giving insights about each scores' relevance to different mappings as well as evaluating different LLMs on the validation task.
 
 Currently, we use a generalist LLM to validate the candidate pairs and justify its decision; but in the future, we hope to fine-tune a Small Language Model or any relevant architecture by using this annotated dataset, that could run quicker and outperform DeepSeek due to its training on the specific task and data.
 
 Lastly, we are going to explore the use of an agentic MCP (Model Context Protocol) server to allow the validation LLM to search for further information online about an entity and solicit a human expertise when it is unsure about a candidate pair. This could benefit the quality of the resulting mapping, countering the lack of information of some resources.
 % Finally, we will also automatize the updates of the mappings as the online vocabularies change, by mapping the added entities to complete our \textit{version manager}.

\section*{Acknowledgments}
\small{\textit{This work has been supported by: the Europlanet 2020 Research Infrastructure (EPN2020-RI) and Europlanet 2024 Research Infrastructure (EPN-2024-RI) projects, which received funding from the European Union's Horizon 2020 research and innovation programme under grant agreement No 654208 and 871149, respectively; the FAIR-IMPACT project, which received funding from the European Commission's Horizon Europe Research and Innovation programme under grant agreement no 101057344; and OPAL cascading grant from the the OSCARS project, which received funding from the European Commission's Horizon Europe Research and Innovation programme under grant agreement no 101129751. The authors also acknowledge support from CNRS and Observatoire de Paris, and especially Stéphane Aicardi and Philippe Hamy, from the Direction Informatique de l'Observatoire (DIO). They also thank Mireille Louys, Emmannuelle Perret and Sébastien Derrière, from CDS (Centre de Données Astronique de Strasbourg, France); and Markus Demleitner (University of Heidelberg, Germany).}}
% %%
%% Define the bibliography file to be used
\bibliography{biblio}

 \appendix

\section{Table of namespaces used in this paper}
%\begin{table}[h!]
  %\label{tab:namespaces}
  \noindent
  \hspace*{-0.5cm}
  \begin{tabular}{l|l|l}
    %\hline
    \textbf{Sematic Artefact name} & \textbf{Namespace} & \textbf{URI} \\
    \hline
    American Astronomical Society%(NS for AAS facilities) 
    & AAS & \url{https://voparis-ns.obspm.fr/rdf/obsfacilities/aas#} \\
    Dublin Core Metadata Initiative (DCMI) & DCTERMS & \url{http://purl.org/dc/terms} \\
    % IVOA Semantics Vocabulary & IVOASEM & \url{https://ivoa.net/rdf/ivoasem/} \\
    WGS84 Geo Positioning & GEO1 & \url{http://www.w3.org/2003/01/geo/wgs84_pos#} \\
    Observation Facilities (our NS) & OBSF & \url{https://voparis-ns.obspm.fr/rdf/obsfacilities#} \\
    Planetary Data System & PDS & \url{https://voparis-ns.obspm.fr/rdf/obsfacilities/pds#} \\
    Resource Description Framework Schema & RDFS & \url{http://www.w3.org/2000/01/rdf-schema#} \\
    Schema & SCHEMA & \url{https://schema.org/} \\
    Semantic Mapping Vocabulary & SEMAPV & \url{https://w3id.org/semapv/vocab/} \\
    Simple Knowledge Organization System & SKOS & \url{http://www.w3.org/2004/02/skos/core#} \\
    A Simple Standard for Sharing Ontology Mappings & SSSOM & \url{https://w3id.org/sssom/} \\ 
  \end{tabular}
  \label{tab:namespaces}
  %\caption{List of namespaces used in this paper.}
%\end{table}
\paragraph{Note} The namespaces of the \url{https://voparis-ns.obspm.fr/rdf/} domain are temporary ones, and are subject to change in further versions.

\section{Extracted data}
\label{appendix:data}
After collecting data from multiple semantic artifacts, we standardize them into a unique ontology.
\begin{verbatim}

aas:european-southern-observatory-1.52m-telescope-at-la-silla-observatory
        a obsf:observatory ;
    dcterms:isPartOf aas:la-silla-observatory ;
    geo1:latitude "-29.2552104"^^xsd:float ;
    geo1:location "South America" ;
    geo1:longitude "-70.739507"^^xsd:float ;
    skos:exactMatch pds:1.52-m-spectrographic-cassegrain-coude-reflector ;
    skos:notation "ESO:1.52m" ;
    skos:prefLabel "European Southern Observatory 1.52m Telescope at La Silla
        Observatory" ;
    obsf:aperture "1.52m" ;
    obsf:waveband wb:infrared,
        wb:optical .


pds:1.52-m-spectrographic-cassegrain-coude-reflector a obsf:telescope ;
    dcterms:description "The 1.52-m spectrographic Cassegrain/Coude reflector is a 1.52 m
    telescope located at -29.255028, 289.267975 at the European Southern Observatory.
    Operational 07/1968+" ;
    dcterms:isPartOf pds:european-southern-observatory,
        pds:european-southern-observatory-la-silla ;
    geo1:latitude "-29.255028"^^xsd:float ;
    geo1:location "Earth" ;
    geo1:longitude "289.267975"^^xsd:float ;
    skos:altLabel "urn:nasa:pds:context:telescope:eso.1m52" ;
    skos:notation "urn:nasa:pds:context:telescope:eso-la_silla.1m52",
        "urn:nasa:pds:context:telescope:eso.1m52" ;
    skos:prefLabel "1.52-m spectrographic Cassegrain/Coude reflector" ;
    schema:url "https://pds.nasa.gov/data/pds4/context-pds4/telescope/eso-la_silla.1m52_1.1.xml" ;
    obsf:altitude "2347" ;
    obsf:aperture "1.52m" ;
    obsf:coordinate_source "Astronomical" ;
\end{verbatim}

    %skos:altLabel "European Southern Observatory (ESO) 1.52m Telescope",
    %    "European Southern Observatory (ESO) 1.52m Telescope at La Silla Observatory" ;
    %schema:Continent "South America" ;
    %schema:address "La Silla Observatory, Acceso Observatorio La Silla, La Higuera, Provincia de Elqui, Coquimbo Region, Chile" ;
    %schema:addressCountry "Chile" ;
    %obsf:location_confidence "0.25"^^xsd:float ;
    %obsf:type_confidence "0.0"^^xsd:float ;
    %obsf:source obsf:aas_list ;

\section{Data enrichment by string parsing}
\label{appendix:alt_labels}
\begin{verbatim}
aas:nasa-0.85m-spitzer-space-telescope a obsf:telescope ;
    geo1:location "Space" ;
    skos:altLabel "NASA 0.85m Spitzer Space Telescope (SST formerly Space Infrared Telescope Facility or SIRTIF) Satellite Mission",
        "NASA 0.85m Spitzer Space Telescope (SST)",
        "SIRTIF",
        "SST", 
        "Space Infrared Telescope Facility",
        "Space Infrared Telescope Facility (SIRTIF)" ;
    skos:notation "Spitzer" ;
    skos:prefLabel "NASA 0.85m Spitzer Space Telescope" ;
    obsf:aperture "0.85m" ;
    obsf:location_confidence "0.5"^^xsd:float ;
    obsf:source obsf:aas_list ;
    obsf:type_confidence "1"^^xsd:float ;
    obsf:waveband wb:infrared .
\end{verbatim}
The alternate labels were extracted from the full label "NASA 0.85m Spitzer Space Telescope (SST formerly Space Infrared Telescope Facility or SIRTIF) Satellite Mission" using string parsing, as well as the telescope aperture (0.85m) which relies on a regular expression.

\section{Scores available for mapping strategies}
 \label{tab:scores}

  \begin{tabular}{|c|c|p{9cm}|}
    \hline
    \textbf{Score name} & \textbf{Type} & \textbf{Description} \\
    \hline
    %\texttt{External identifier} & Accept & Entities that are linked by an external identifier. \\
    \texttt{Label match} & Accept & Entities match if any or their aliases are equal. \\
    \texttt{Identifier} & Reject & Eliminate candidate pairs when one of their identifiers mismatch (\href{https://www.wikidata.org/wiki/Property:P2956}{NAIF ID}), \href{https://www.wikidata.org/wiki/Property:P247}{COSPAR ID}, \href{https://www.wikidata.org/wiki/Property:P8913}{NSSDCA ID}). \\
    \texttt{Distance limit} & Reject & Eliminate ground-based facilities if their geodesic distance > 4km. \\
    \texttt{Date mismatch} & Reject & Eliminate launch, start and/or end date mismatch. \\
    \texttt{Aperture mismatch} & Reject & Eliminate facilities with a different lens aperture. \\
    \texttt{Acronym probability} & Surface & Probability of a label to be the acronym of another label. \\
    \texttt{Levenshtein similarity} & Surface & Edit distance between labels. Can detect typos or small naming divergences. \\
    \texttt{Digits Match} & Surface & Digits match ratio between all fields. \\
    \texttt{TF-IDF} & Semantic & Encode each word of the entity's textual fields and compute a cosine similarity. BoW approach, not multilingual, not synonym-aware. \\
    \texttt{Sentence transformer} & Semantic & Encode semantic fields' sentences and compute a cosine similarity. Multilingual, synonym-aware. \\
    \texttt{LLM embeddings} & Semantic & Cosine similarity on embeddings produced by an LLM. Multilingual, synonym-aware. \\
    \hline
  \end{tabular}
%\end{table}

\section{Mapping strategy configuration file}
\label{appendix:mapping_strat}
This strategy does not use neural network scores (sentence transformer or LLM embeddings) in order to run faster.
\begin{verbatim}
iaumpc, wikidata[spacecraft]: label_match, identifier, levenshtein, tfidf, digit
iaumpc, wikidata[all,-spacecraft]: label_match, identifier, distance, type, levenshtein, tfidf
spase, nssdc: type, label_match, identifier, date, levenshtein, tfidf, digit
spase, iaumpc: type, label_match, identifier, distance, date, levenshtein, tfidf, digit
pds, wikidata: label_match, distance, date, levenshtein, tfidf, digit
pds, aas: distance, type, date, aperture, label_match, levenshtein, tfidf, digit
imcce, naif[spacecraft]: label_match, date, levenshtein, tfidf, digit
\end{verbatim}

\section{SSSOM ontology}
\label{appendix:sssom}

If the candidate pair is accepted by a label match, it does not need to be reviewed by an LLM. We save a simple Mapping entity in the SSSOM ontology:
\begin{verbatim}
obsf:b7cee265-8355-4d4c-a05d-dc22cd63592c a sssom:Mapping ;
    obsf:label_match "1"^^xsd:float ;
    sssom:mapping_date "2025-07-23T11:12:18.890248"^^xsd:dateTimeStamp ;
    sssom:mapping_tool "FacilityList/merge.py" ;
    sssom:object_id aas:observatorio-del-teide> ;
    sssom:predicate_id skos:exactMatch ;
    sssom:similarity_measure "label_match" ;
    sssom:similarity_score "1"^^xsd:float ;
    sssom:subject_id pds:observatorio-del-teide> .
\end{verbatim}

If the validation is done by an LLM, we save the scores details and the reviewing metadata in the SSSOM Ontology:

\begin{verbatim}
obsf:10d6f11e-4552-4343-a986-295e206543ed a sssom:Mapping ;
    rdfs:comment "both entities refer to a 1.52m telescope located at la silla
    observatory in chile, operated by the european southern observatory (eso).
    they share identical attributes such as aperture size (1.52m), location
    (south america, chile), and waveband capabilities (infrared, optical).
    the slight differences in namingc onventions (\"european southern
    observatory 1.52m telescope at la silla observatory\" vs. \"1.52-m
    spectrographic cassegrain/coude reflector\") do not indicate distinct
    entities but rather different descriptive labels for the same telescope.
    therefore, they are the same entity." ;
    obsf:levenshtein_similarity "0.4444444444444444"^^xsd:float ;
    obsf:weighted_sum "0.3277075222694277"^^xsd:float ;
    obsf:tfidf_cosine_similarity "0.2926864456169227"^^xsd:float ;
    sssom:justification semapv:LexicalMatching ;
    sssom:mapping_date "2025-07-23T20:08:51.577381"^^xsd:dateTimeStamp ;
    sssom:mapping_tool "FacilityList/merge.py" ;
    sssom:object_id aas:european-southern-observatory-1.52m-telescope-
                        at-la-silla-observatory> ;
    sssom:predicate_id skos:exactMatch ;
    sssom:reviewer_label "deepseek-v3:671b-q4_K_M" ;
    sssom:similarity_measure "weighted_sum" ;
    sssom:similarity_score "0.3277075222694277"^^xsd:float ;
    sssom:subject_id pds:1.52-m-spectrographic-cassegrain-coude-reflector> .

\end{verbatim}

\section{Sample of the generated JSON for the name resolver}
\label{appendix:json}
\begin{verbatim}
{...
  "3.6-m-equatorial-cassegrain-coude-reflector": [
    "3.6-m equatorial Cassegrain/Coude reflector",
    "urn:nasa:pds:context:telescope:eso.3m6"
  ],
  "isee-magnetometer-nain-station": [
    "ISEE Magnetometer Nain station",
    "spase://IUGONET/Observatory/ISEE/Induction/NAI"
  ],
  "cosmos-1221": [
    "1980-090A",
    "12058",
    "COSMOS 1221"
  ],
  ...}
\end{verbatim}

 % The sources for the ceur-art style are available via
 % \begin{itemize}
 % \item \href{https://github.com/yamadharma/ceurart}{GitHub},
 % \item \href{https://www.overleaf.com/project/5e76702c4acae70001d3bc87}{Overleaf},
 % \item
 %   \href{https://www.overleaf.com/latex/templates/template-for-submissions-to-ceur-workshop-proceedings-ceur-ws-dot-org/pkfscdkgkhcq}{Overleaf
 %     template}.
 % \end{itemize}

\end{document}